\documentclass[nohyperref]{article}

\usepackage{microtype}
\usepackage{graphicx}
\usepackage{subfigure}
\usepackage{booktabs} % for professional tables
\usepackage{amsfonts,amssymb}
  \usepackage{mathrsfs}
\usepackage{hyperref}

\usepackage[accepted]{icml2022}

\usepackage{amsmath}
\usepackage{amssymb}
\usepackage{mathtools}
\usepackage{amsthm}
\usepackage{multirow}
\usepackage[capitalize,noabbrev]{cleveref}

\theoremstyle{plain}
\newtheorem{theorem}{Theorem}[section]

\newtheorem{lemma}[theorem]{Lemma}

\theoremstyle{definition}
\newtheorem{definition}[theorem]{Definition}

\theoremstyle{remark}

\usepackage[textsize=tiny]{todonotes}

\icmltitlerunning{Interventional Contrastive Learning with Meta Semantic Regularizer}

\begin{document}

\twocolumn[
\icmltitle{Interventional Contrastive Learning with Meta Semantic Regularizer}



\icmlsetsymbol{equal}{*}

\begin{icmlauthorlist}
\icmlauthor{Wenwen Qiang}{equal,yyy,comp,compp}
\icmlauthor{Jiangmeng Li}{equal,yyy,comp,compp}
\icmlauthor{Changwen Zheng}{yyy,compp}
\icmlauthor{Bing Su}{sch,bjk}
\icmlauthor{Hui Xiong}{yyyy,yyyyy}
\end{icmlauthorlist}
\icmlaffiliation{yyy}{Science \& Technology on Integrated Information System Laboratory, Institute of Software Chinese Academy of Sciences, Beijing, China}
\icmlaffiliation{comp}{University of Chinese Academy of Sciences, Beijing, China}
\icmlaffiliation{compp}{Southern Marine Science and Engineering Guangdong Laboratory (Guangzhou), Guangdong, China.}
\icmlaffiliation{sch}{Gaoling School of Artificial Intelligence, Renmin University of China, Beijing, China}
\icmlaffiliation{bjk}{Beijing Key Laboratory of Big Data Management and Analysis Methods, Beijing, China}
\icmlaffiliation{yyyy}{Thrust of Artificial Intelligence, The Hong Kong University of Science and Technology (Guangzhou), Guangzhou, China}
\icmlaffiliation{yyyyy}{Department of Computer Science \& Engineering, The Hong Kong University of Science and Technology, Hong Kong SAR, China}
\icmlcorrespondingauthor{Bing Su}{subingats@gmail.com}
%\icmlcorrespondingauthor{Firstname2 Lastname2}{first2.last2@www.uk}

\icmlkeywords{Contrastive Learning, Self-Supervised
Learning, Structural Causal Mode, Backdoor Adjustment, Classification}

\vskip 0.3in
]



\printAffiliationsAndNotice{\icmlEqualContribution} % otherwise use the standard text.

\begin{abstract}
Contrastive learning (CL)-based self-supervised learning models learn visual representations in a pairwise manner. Although the prevailing CL model has achieved great progress, in this paper, we uncover an ever-overlooked phenomenon: When the CL model is trained with full images, the performance tested in full images is better than that in foreground areas; when the CL model is trained with foreground areas, the performance tested in full images is worse than that in foreground areas. This observation reveals that backgrounds in images may interfere with the model learning semantic information and their influence has not been fully eliminated. To tackle this issue, we build a Structural Causal Model (SCM) to model the background as a confounder. We propose a backdoor adjustment-based regularization method, namely \textit{Interventional Contrastive Learning with Meta Semantic Regularizer} (ICL-MSR), to perform causal intervention towards the proposed SCM. ICL-MSR can be incorporated into any existing CL methods to alleviate background distractions from representation learning. Theoretically, we prove that ICL-MSR achieves a tighter error bound. Empirically, our experiments on multiple benchmark datasets demonstrate that ICL-MSR is able to improve the performances of different state-of-the-art CL methods.
\end{abstract}

\section{Introduction}
\label{submission}
Learning robust and generic representations without human annotation is a long-standing and important topic in machine learning. Contrastive learning (CL)-based self-supervised learning, an innovative unsupervised representation learning (SSL) method, has recently demonstrated superiority in computer vision tasks such as classification \cite{i1}, object identification \cite{i2}, and transfer learning \cite{i3}. The success of CL is partly due to its instance-based learning paradigm, e.g., CL assumes that each sample in the training dataset as a distinct class. This paradigm can be applied to any type of data to capture common semantic information applicable to different tasks.

In general, for a sample $X$ in a mini-batch of training data, two augmented samples $X^1$ and $X^2$ are generated by performing random augmentation transformations $\mathcal{T}$ to $X$, i.e., $\left\{ {X^1, X^2} \right\} = \mathcal{T}(X)$. Then, using one of two augmented samples as the anchor, most existing CL frameworks treat the remaining augmented sample as a positive sample and the augmented samples generated by the other samples in the mini-batch as negative samples. The contrastive loss \cite{i1} is used to train the feature extractor. According to the instance-based learning paradigm, minimizing contrastive loss entails pulling the positive sample closer to the anchor and pushing the negative samples further away from the anchor in the learnt feature space \cite{o2}. Also, the contrastive loss can be considered as a way to assess the mutual information between the positive sample and the anchor from an information theoretic standpoint \cite{o1}. The high similarity or mutual information between the positive sample and the anchor should be due to shared semantic or foreground-related information. Observations from several toy experiments, on the other hand, contradict this.

\begin{figure}[t]
	\centering
	\includegraphics[width=0.48\textwidth]{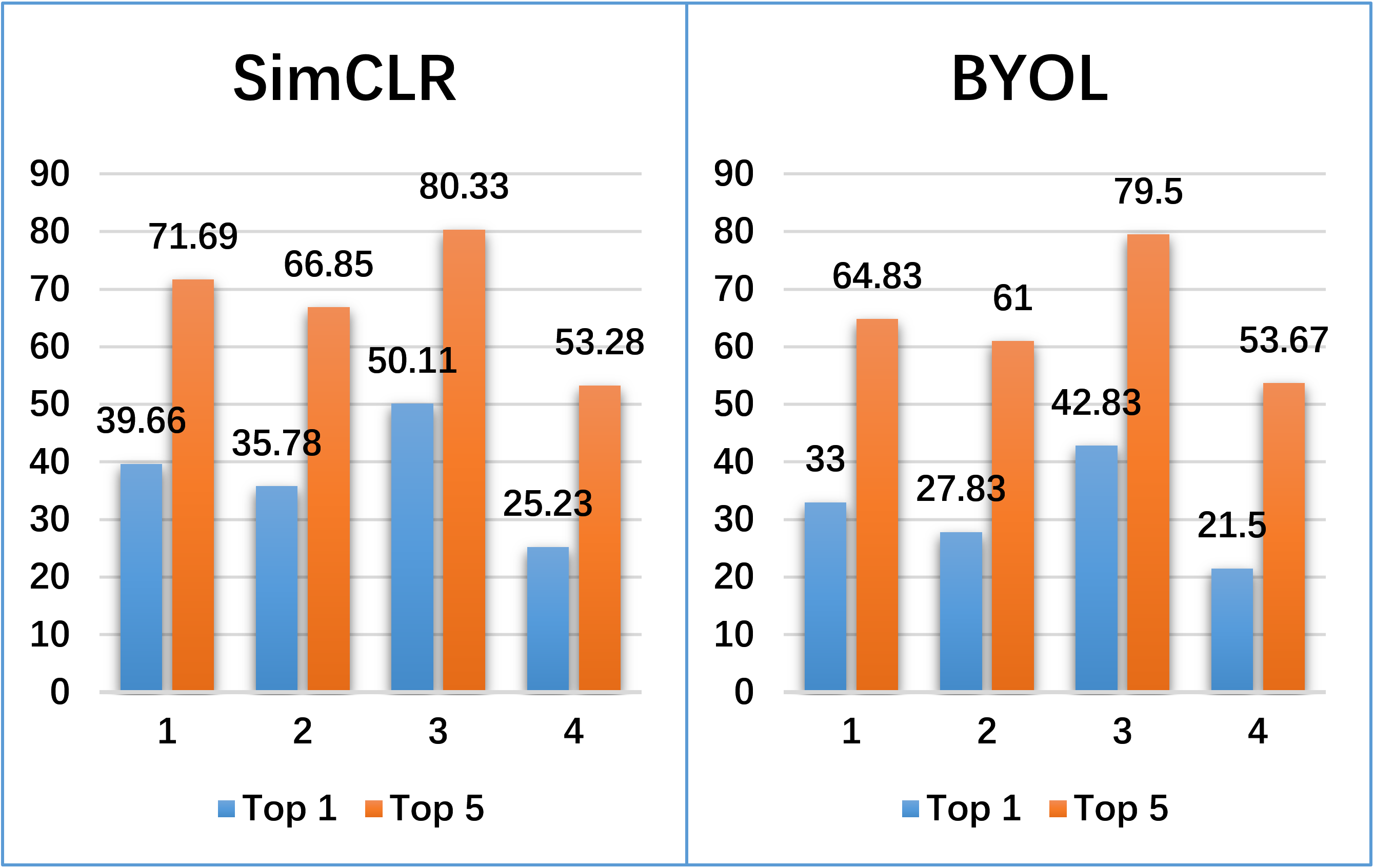}
	\vskip -0.1in
	\caption{The experimental results for two CL models. "1" represents training and testing on full images, "2" represents training on full images and testing on foreground images, "3" represents training and testing on foreground images, and "4" represents training on foreground images and testing on full images.}
	\label{toy}
	\vskip -0.2in
\end{figure}

To be more specific, we run the toy experiments on the COCO dataset \cite{lin2014microsoft} with four different experimental settings: 1) training and testing the CL model on full images; 2) training and testing the CL model on full images and foreground images, respectively; 3) training and testing the CL model on foreground images; and 4) training and testing the CL model on foreground images and full images, respectively. SimCLR \cite{i1} and BYOL \cite{i2}, two CL models, were chosen as baselines. Figure \ref{toy} shows an often-overlooked characteristic in the current CL model: Comparing the results produced under settings 1) and 2), where the model is trained on full images, the performance evaluated on full images is clearly superior than the performance tested on foreground images. In addition, when comparing the results obtained under configurations 1) and 4), the model trained with full images outperforms the model trained with foreground images when both are tested on full images. That is, when the backdrop is removed from the full image during training or testing, the performance of the two CL models suffers. This discovery suggests that background-related information can influence the CL models' learning process. However, comparing the results obtained under settings 3) and 4), we discover that when the model is trained with foreground images, the performance tested in foreground images is considerably better than the performance tested in full images. In addition, when all variables are considered, we find that training and testing with only foreground images produce the best results. We can confidently conclude that background-related information degrades the performance of the CL models based on this. As we can see, the two conclusions are mutually exclusive.

A plausible explanation is that a feature extractor trained on full images extracts background-dependent semantic features. During the test phase, because the full image contains both foreground and background parts, besides the foreground parts, the background part also plays a certain role in promoting the classification. CL, on the other hand, strives to be adaptable to a variety of downstream tasks, such as object detection, object segmentation, and so on. Only foreground-related semantic information can ensure the robustness of the learned features to various tasks. This is in accordance with the second observation. For this purpose, we develop a Structural Causal Model (SCM) to describe the causal relationships between semantic information, positive sample, and anchor in this paper. We can represent the background as a confounder based on this, which fits with the explanation given above. Then, to execute causal intervention towards the proposed SCM, we present a backdoor adjustment-based regularization approach called Interventional Contrastive Learning with Meta Semantic Regularizer (ICL-MSR). To eliminate background distractions from representation learning, ICL-MSR can be simply implemented into most existing CL approaches. We show that ICL-MSR achieves a tighter error bound than CL methods that merely minimize the contrastive loss. Our experiments on multiple benchmark datasets show that ICL-MSR can improve the performance of the state-of-the-art CL-based self-supervised learning approaches. The following is a list of our major contributions:
\begin{itemize}
\item We discover a paradox: under different setting, background information can both improve and prevent performance of the learned feature representations from improving.

\item To capture the causal links between semantic information, positive sample, and anchor, we establish a Structural Causal Model (SCM). We can simply deduce that the background is effectively a confounder that produces misleading correlations between the positive sample and the anchor based on this.

\item We propose a new method called Interventional Contrastive Learning with Meta Semantic Regularizer (ICL-MSR) by implementing backdoor adjustments to the planned SCM. 

\item  We provide theoretical guarantee on the error bound and empirical evaluations to demonstrate that ICL-MSR can improve the performances of different state-of-the-art CL methods.

\end{itemize}

\section{Related Works}
Self-supervised learning, such as contrastive learning (CL), aims to learn a generalized feature extractor that can be well applied to downstream tasks. The objective of contrastive learning is mainly based on the InfoNCE loss. It is first proposed in \cite{o1} and can be seen as a lower bound of the mutual information between the feature and the context. SimCLR \cite{i1, chen2020big, bt2, wen2021toward} extends the InfoNCE loss to maximize the similarity between two different data augmentations. To better prompt the performance of contrastive learning, InfoMin \cite{tian2020makes} proposes a set of stronger augmentations that reduce the mutual information between views while keeping task-relevant information intact. AlignUniform \cite{o2} relates the contrastive loss to two critical properties, including alignment and uniformity, to form a new objective. The CL model has shown its superiority in many vision tasks. However, there are still some challenges worth mentioning. 

Firstly, CL is sensitive to batch size. To solve this problem, MoCo \cite{ he2020momentum, chen2020improved, chen2021empirical} increases the number of negative examples by using a memory bank. Secondly, some negative samples may contain similar semantic information as the positive sample. To tackle this issue, SwAV \cite{caron2020unsupervised} and PCL \cite{li2020prototypical} learn good-quality negative samples by introducing clustering methods in the training process, thereby reducing the number of negative samples. BYOL \cite{i2} proposes learning feature representation without negative samples. However, this also brings a new challenge: degenerate solutions. So, BYOL proposes learning feature representations with a structurally asymmetric feature extractor and adopting the moving average as an optimization method for model parameters. Then, DirectPred \cite{ tian2021understanding} provides theoretical analysis to verify the effectiveness of BYOL. At the same time, a series of effective works such as SimSiam \cite{ chen2021exploring}, Barlow Twins \cite{i3}, W-MSE \cite{ ermolov2021whitening}, and SSL-HSIC \cite{bt2} are proposed to avoid degenerate solutions. Among them, W-MSE and Barlow Twin do not require asymmetric networks and are conceptually simpler. In addition to the improvement of the model, the contrastive learning theory is also attracting increasing attention. Some works \cite{nozawa2021understanding, arora2019theoretical} provide a bound on the CL. RELIC \cite{mitrovic2020representation} gives a causal explanation for the objective of CL.

The goal of this paper is to explore the impact of background information on the representation learning process in contrastive learning. Our proposed ICL-MSR can be incorporated into most existing CL methods to alleviate background distractions. Also, there are three differences between ICL-MSR and Causal3DIdent \cite{von2021self}. The first is that ICL-MSR is motivated by causal intervention, while Causal3DIdent is motivated by counterfactual. The second is that ICL-MSR is mainly concerned with the objective function of contrastive learning and regards the background-dependent semantic features as confounding factors, while Causal3DIdent focuses on data augmentation and regards it as counterfactual under soft style intervention. The third is that ICL-MSR implements backdoor adjustment, and Causal3DIdent can be a part of ICL-MSR.

\section{Problem Formulations}
\subsection{Contrastive Learning}
In this paper, we mainly focus on CL-based representation learning approaches. The primary goal of the CL methods is to learn a generic feature extractor $f$, which projects the sample from the original input space to a latent space for extracting intrinsic features.

Formally, we first randomly sample a minibatch of training data and denote it as ${X_{tr}} = \left\{ {{X_i}} \right\}_{i = 1}^N$, where $X_i$ represents the $i$-th sample, and $N$ represents the number of samples in the minibatch. We perform stochastic data augmentation (e.g., random crop) to transform each sample $X_i$ into two augmented views $X_i^1$ and $X_i^2$. Since there are $N$ samples in ${X_{tr}}$, we finally obtain $2N$ augmented samples denoted as ${X_{tr}^{aug}} = {\left\{ {X_i^1,X_i^2} \right\}_{i = 1}^N}$. Then, we feed all training samples into the feature extractor $f$ to get their feature representations, i.e., $Z_i^j = f(X_i^j),i \in \left\{ {1,...,N} \right\},j \in \left\{ {1,2} \right\}$. The general objective of CL is formulated as:
\begin{equation}
\label{loss:cl}
{L_{ct}} = \sum\limits_{i = 1}^N {\sum\limits_{j = 1}^2 { - \log \frac{{\exp \left( {\frac{{{\mathop{\rm sim}\nolimits} \left( {Z_i^j,Z_i^{3 - j}} \right)}}{\tau }} \right)}}{{\sum\limits_{k = 1, }^N {\sum\limits_{l = 1,l \ne j}^2 {\exp \left( {\frac{{{\mathop{\rm sim}\nolimits} \left( {Z_i^j,Z_k^l} \right)}}{\tau }} \right)} } }}} } 
\end{equation}
where $\tau$ represents the temperature hyper-parameter and ${\mathop{\rm sim}\nolimits} \left( {u,v} \right) = {{{u^T}v} \mathord{\left/
 {\vphantom {{{u^T}v} {\left\| u \right\|\left\| v \right\|}}} \right.
 \kern-\nulldelimiterspace} {\left\| u \right\|\left\| v \right\|}}$ denotes the dot product between $l_2$ normalized $u$ and $v$ (i.e.,
the cosine similarity). For a sample $X_i^j$ randomly selected from the dataset ${X_{tr}^{aug}}$, CL regards it as the anchor, the pair $\{X_i^j,X_i^{3-j}\}$ as positive, the pairs $\{ X_i^j,X_k^l\} _{k = 1,k \ne i,l = 1}^{k = N,l = 2}$ as negatives. The loss ${L_{ct}}$ in objective (\ref{loss:cl}) is computed across all positive pairs.

\begin{figure}[t]
	\centering
	\includegraphics[width=0.45\textwidth]{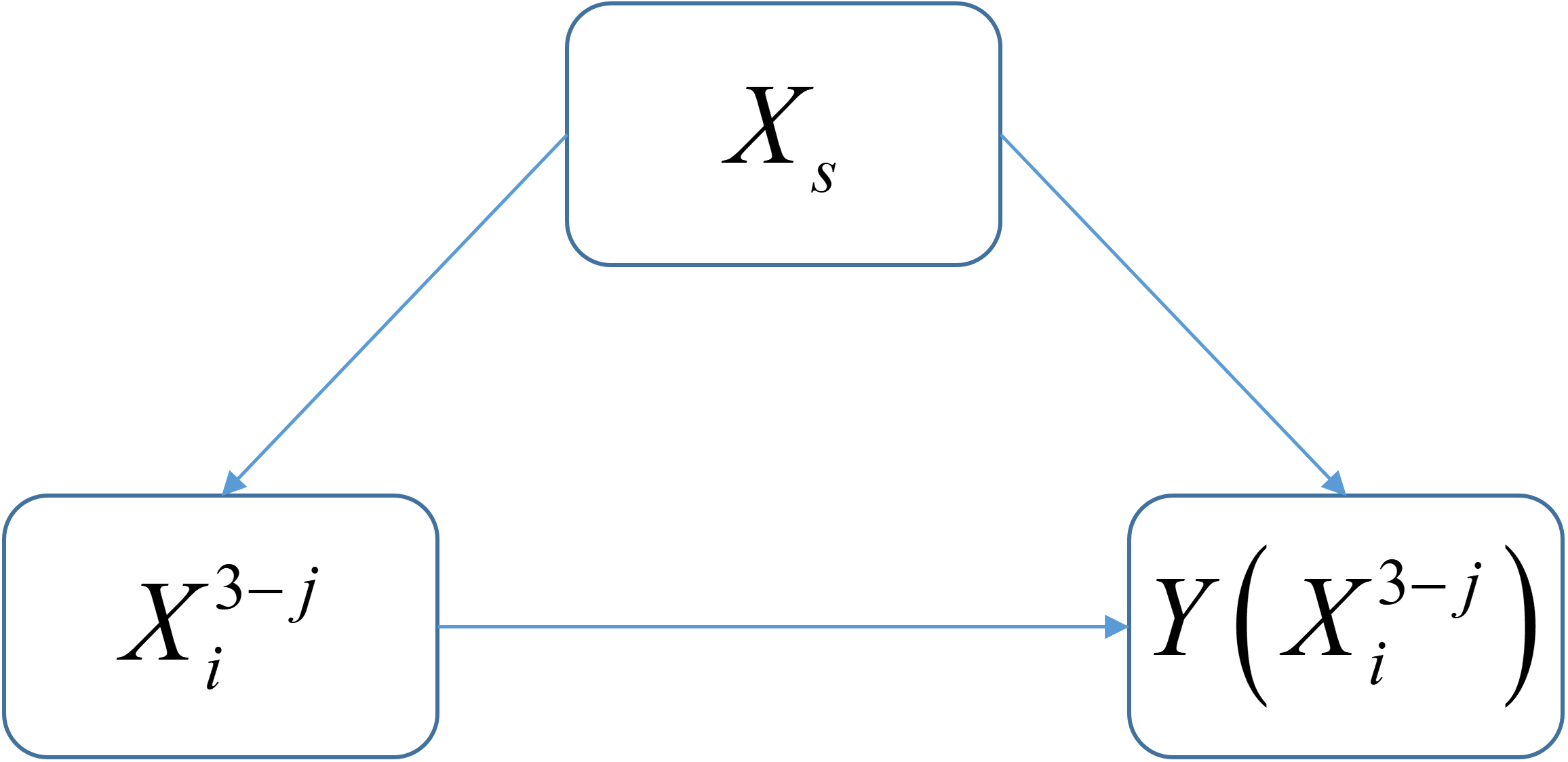}
	\vskip -0.1in
	\caption{The proposed SCM between semantic information $X_s$, positive sample $X^{3-j}_i$, and anchor (or label) $Y({X^{3-j}_i})$.}
	\label{SCM}
	\vskip -0.15in
\end{figure}

\begin{figure*}[t]
	\centering
	\includegraphics[width=0.95\textwidth]{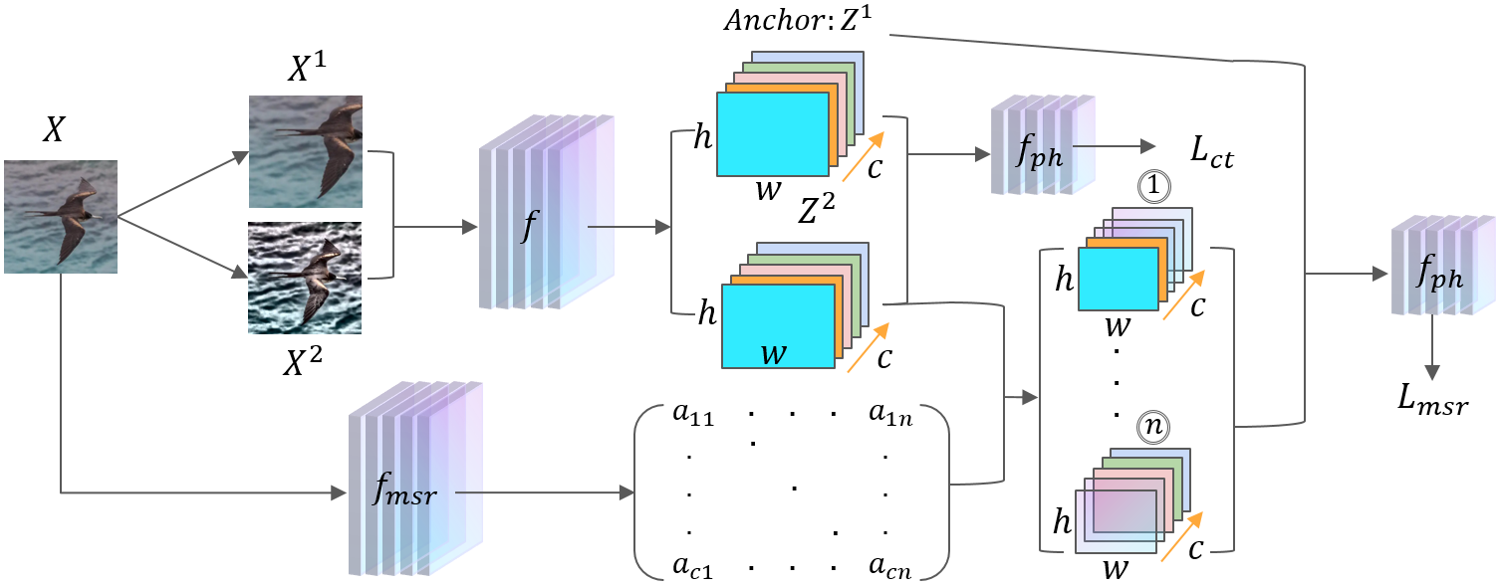}
	\vskip -0.1in
	\caption{The framework of the proposed ICL-MSR.}
	\label{frm}
	\vskip -0.15in
\end{figure*}

\subsection{Structural Causal Model}

Minimizing the objective (\ref{loss:cl}) is to make the sample $X_i^{3-j}$ in the positive pair close to the anchor and the sample $X_k^l$ in the negative pairs far away from the anchor. From this perspective, the anchor can be seen as the label, then minimizing the objective (\ref{loss:cl}) equals to predict $X_i^{3-j}$ to the label $X_i^j$. Then, the SCM implicated in CL can be formalized as Figure \ref{SCM}. The nodes in SCM represent the abstract data variables and the directed edges represent the (functional) causality, e.g., $
X_i^{3 - j} \to Y(X_i^{3 - j})$ represents that $X_i^{3 - j}$ is the cause and $Y(X_i^{3 - j})$ is the effect. In the following, we will describe the proposed SCM and the rationale behind its construction in detail at a high level. Please refer to Section \ref{se4} for the detailed functional implementations.

$X_i^{3 - j} \to Y(X_i^{3 - j})$. $Y(X_i^{3 - j})$ denotes the corresponding label of $X_i^{3 - j}$. As mentioned above, the label equals to the anchor. Thus, we have $Y(X_i^{3 - j}) = X_i^j$. This link assumes that $X_i^{3 - j}$ should be similar with the anchor $X_i^j$.

$X_i^{3 - j} \leftarrow {X_s} \to Y( {X_i^{3 - j}} )$. ${X_s}$ represents the semantic information and can be regarded as the convolution kernel of the feature extractor $f$. Therefore, the link $X_i^{3 - j} \leftarrow {X_s}$ and ${X_s} \to Y( {X_i^{3 - j}} )$ assume that the feature representations of $X_i^{3 - j}$ and $Y( {X_i^{3 - j}} )$ in the latent space is extracted by using $f$, and each feature representation channel corresponds to a semantic information.

An ideal contrastive learning model should capture the true causality between $X_i^{3 - j}$ and $Y(X_i^{3 - j})$ and can generalize to unseen samples well. For example, we expect the label prediction of $Y(X_i^{3 - j})$ to be caused by the foreground feature, not the background information. However, from the proposed SCM, the increased likelihood of $Y(X_i^{3 - j})$ given $X_i^{3 - j}$ is not only due to $X_i^{3 - j} \to Y(X_i^{3 - j})$, but also the spurious correlation via $X_s \to X_i^{3 - j} \to Y(X_i^{3 - j})$, e.g., the background of $X_i^{3 - j}$ generates the background feature, which provides useful context for predicting the anchor, this corresponds to our first observation in Figure \ref{toy}: when the model is trained on full images, the performance evaluated on full images is clearly superior to the performance tested on foreground images. Therefore, to pursue the true causality between $X_i^{3 - j}$ and $Y(X_i^{3 - j})$, we need to use the causal intervention $P( {Y(X_i^{3 - j})| {do( {X_i^{3 - j}} )} } )$ instead of the $P( {Y(X_i^{3 - j})| {X_i^{3 - j}} } )$ to measure the causal relation.

\subsection{Causal Intervention via Backdoor Adjustment}
Before we introduce the backdoor adjustment, we first give an intuition of why the feature extractor $f$ trained by previous contrastive learning models extracts background-dependent semantic features.

The fact that the size of the positive sample is too tiny, i.e. only one, could explain this problem. As shown in objective (\ref{loss:cl}), randomly given an anchor, there is only one positive sample, but $2N-2$ negative samples. Also, there are some samples (false negative samples) in negative pairs that are with the same foreground as the positive sample. Due to the randomness of the sampling process, the background similarity between false negative samples and anchor is lower than the foreground similarity between them. However, the anchor and the sample in the positive pair are generated by the same original image, so that the foreground and background between the two positive samples are similar. When minimizing the objective (\ref{loss:cl}), the shared foreground and background between the positive pair will likely together drag the positive sample towards the anchor if there are too few positive samples. Observation is also carried out on false negative samples that have semantically equivalent information to the anchor \cite{tian2020contrastive}. As a result, shifting the false negative samples further from the anchor will mainly focus on lowering the foreground similarity. In other word, this in turn may promote "drag" operation to pay greater attention to the backdrop to some extent. As a result, the role of the background is enhanced. This can lead to the previous contrastive learning models extracting background-dependent semantic features.

Above, we have shown that the feature extractor $f$ can extract background-dependent semantic features and analyze how these semantic features affect the training process of the contrastive learning model. Below, we will propose to use the backdoor adjustment \cite{glymour2016causal} to eliminate the interference of the background-dependent semantics to achieve $P( {Y(X_i^{3 - j})| {do( {X_i^{3 - j}} )} } )$.

Specifically, the backdoor adjustment assumes that we can observe and stratify the confounder. In the proposed SCM, the confounder is contained in $X_s$, we can layer it into different semantic features, e.g., $X_s = \{Z_s^i\}_{i=1}^{i=n}$, where $Z_s^i$ represents a stratification
of semantic features. Formally, the backdoor adjustment for the proposed SCM is presented as: 
\begin{equation}
\label{bd}
\begin{array}{l}
P( {Y( {X_i^{3 - j}} )| {do( {X_i^{3 - j}} )} } )\\
 \;\;\;\;\;\;\;\;\;\;= \sum\limits_{i = 1}^n {P( {Y( {X_i^{3 - j}} )| {X_i^{3 - j},Z_s^i} } )P( {Z_s^i} )} 
\end{array}
\end{equation}
where $P( {Y( {X_i^{3 - j}} )| {do( {X_i^{3 - j}} )} } )$ represents the true causality between ${Y( {X_i^{3 - j}} )}$ and $X_i^{3 - j}$. See appendix \ref{bdbd} for detailed derivation of equation (\ref{bd}).

\section{Methodology}\label{se4}
\subsection{Meta Semantic Regularizer}
In this subsection, we present the implementation of the backdoor adjustment during the training phase. As shown in equation (\ref{bd}), firstly, we need to give detailed functional implementations for $Z_s^i$. 

Without loss of generality, we denote the dimension of the output $Z_i^j$ of the feature extractor $f$ as $w \times h \times c$, where $w$ is the width, $h$ is the height, and $c$ is the number of feature channels. To be more specific, we denote $Z_i^j$ as $Z_i^j = [ {Z_{i,1}^j,...,Z_{i,c}^j} ]$, where $Z_{i,r}^j \in {\mathbb{R}^{w \times h}}$ represents a feature map of $Z_i^j$, $r \in \left\{ {1,...,c} \right\}$. Note that for any pre-trained CNN-based feature extractor, each channel corresponds to a kind of semantic information or visual concepts \cite{z76, z82}. However, it is difficult to encode one visual concept by a single channel. So, this motivates us to find a weight vector for each semantic information. Our idea is that each semantic information corresponds to one subset
of channels. So the weights for this related subset of channels should be large, and the weights for channels that are outside of the subset should be small. 

Given a weight vector ${a_t} = {\left[ {{a_{1,t}},...,{a_{c,t}}} \right]^{\mathop{\rm T}\nolimits} }$, we represent the functional implementations of the $Z_s^t$ as $Z_s^t = a_t$, and ${P( {Z_s^t} )}  = {1 \mathord{\left/
 {\vphantom {1 n}} \right.
 \kern-\nulldelimiterspace} n}$. Then, we represent the functional implementations of the ${P( {Y( {X_i^{3 - j}} )| {X_i^{3 - j},Z_s^t} } )}$ as 
\begin{equation}
\begin{array}{l}
P( {Y( {X_i^{3 - j}} )| {X_i^{3 - j},Z_s^t} } ) = \\
\frac{{\exp \left( {\frac{{{\mathop{\rm sim}\nolimits} \left( {Z_i^j,{a_t} \odot Z_i^{3 - j}} \right)}}{\tau }} \right)}}{{\exp \left( {\frac{{{\mathop{\rm sim}\nolimits} \left( {Z_i^j,{a_t} \odot Z_i^{3 - j}} \right)}}{\tau }} \right) + \sum\limits_{k = 1,\hfill\atop
k \ne i\hfill}^N {\sum\limits_{l = 1,\hfill\atop
l \ne j\hfill}^2 {\exp \left( {\frac{{{\mathop{\rm sim}\nolimits} \left( {Z_i^j,Z_k^l} \right)}}{\tau }} \right)} } }},
\end{array}
\end{equation}
where ${a_t} \odot Z_i^{3-j} = [ {{a_{1,t}} \cdot Z_{i,1}^{3-j},...,{a_{c,t}} \cdot Z_{i,c}^{3-j}} ]$. As a result, the overall backdoor adjustment is presented as:
\begin{equation}
\begin{array}{l}
P( {Y( {X_i^{3 - j}} )| {do( {X_i^{3 - j}} )} } ) = \\
\sum\limits_{t = 1}^n {\frac{{\exp \left( {\frac{{{\mathop{\rm sim}\nolimits} \left( {Z_i^j,{a_t} \odot Z_i^{3 - j}} \right)}}{\tau }} \right) \times \frac{1}{n}}}{{\exp \left( {\frac{{{\mathop{\rm sim}\nolimits} \left( {Z_i^j,{a_t} \odot Z_i^{3 - j}} \right)}}{\tau }} \right) + \sum\limits_{k = 1,\hfill\atop
k \ne i\hfill}^N {\sum\limits_{l = 1,\hfill\atop
l \ne j\hfill}^2 {\exp \left( {\frac{{{\mathop{\rm sim}\nolimits} \left( {Z_i^j,Z_k^l} \right)}}{\tau }} \right)} } }}}.
\end{array}
\end{equation}

The proposed meta semantic regularizer can be thought of as a learnable module $f_{msr}$, which is to generate the semantically relevant weight matrix $A_s$ and is implemented by convolutional neural network, where ${A_s} = \left[ {{a_1},...,{a_n}} \right]$. Specifically, for a input sample $X$, we first obtain two augmentated samples $\left\{ {{X^1},{X^2}} \right\}$ by feeding $X$ to a stochastic data augmentation module. Then we feed $X$ to the module $f_{msr}$ to obtain the weight matrix $A_s$, and the two augmented samples $\left\{ {{X^1},{X^2}} \right\}$ share only one weight matrix $A_s$. 

\subsection{Model Objectives}
To this end, we introduce the objective of the proposed interventional contrastive learning with meta semantic regularizer (ICL-MSR). The whole learning framework of ICL-MSR is shown in Figure \ref{frm}, and the training process is shown in Appendix. ICL-MSR consists of three modules, including the feature extractor $f$, the meta semantic regularizer $f_{msr}$, and the projective head $f_{ph}$. The meta semantic regularizer is trained alongside the feature extractor, with two stages per epoch. In the first stage, $f$ and $f_{ph}$ are learned using the two augmented training dataset $X_{tr}^{aug}$ and the semantically relevant weight matrix $A_s$. In the second stage, $f_{msr}$ is updated by computing its gradients with respect to the contrastive loss. We train both modules in an iterative manner until convergence.

In the first stage of each epoch, the parameters of $f$ and $f_{ph}$ are updated by minimizing the objective $L_{to}$, which can be presented as:
\begin{equation}
    \mathop {\min }\limits_{f,{f_{ph}}} {L_{to}} = {L_{ct}} + \lambda {L_{msr}},
\end{equation}
where $L_{ct}$ is shown in the objective (\ref{loss:cl}), $\lambda$ is the hyper-parameter, and  $L_{msr}$ is presented as:
\begin{equation}
    {L_{msr}} = \sum\limits_{i = 1}^N {\sum\limits_{j = 1}^2 { - \log P( {Y( {X_i^{3 - j}} )| {do( {X_i^{3 - j}} )} } )} }.
\end{equation}

\begin{table*}[t]
	\small
	\renewcommand\arraystretch{1.22}
	\vskip 0.in
	\caption{Classification accuracy (top 1) of a linear classifier and a 5-nearest neighbors classifier for methods and datasets
with the ResNet-18 feature extractor.}
	\vskip 0.15in
	\label{tab:1}
	\setlength{\tabcolsep}{10pt}
	\begin{center}
		\begin{tabular}{lcccccccc}
			\toprule
			\multirow{2}{*}{Methods} & \multicolumn{2}{c}{CIFAR-10} & \multicolumn{2}{c}{CIFAR-100} & \multicolumn{2}{c}{STL-10} & \multicolumn{2}{c}{Tiny ImageNet} \\ 
			\cline{2-9}
			& linear & 5-nn & linear & 5-nn & linear & 5-nn & linear & 5-nn \\
			\midrule
			\text{SimCLR \cite{i1}} & 91.80 & 88.42 & 66.83 & 56.56 & 90.51 & 85.68 & 48.84 & 32.86 \\
			\text{BYOL \cite{i2}} & 91.73 & 89.45 & 66.60 & 56.82 & 91.99 & 88.64 & 51.00 & 36.24 \\
			\text{W-MSE \cite{ermolov2021whitening}} & 91.99 & 89.87 & 67.64 & 56.45 & 91.75 & 88.59 & 49.22 & 35.44 \\
			\text{ReSSL \cite{zheng2021ressl}} & 90.20 & 88.26 & 63.79 & 53.72 & 88.25 & 86.33 & 46.60 & 32.39\\
			\text{LMCL \cite{chen2021large}} & 91.91 & 88.52 & 67.01 & 56.86 & 90.87 & 85.91 & 49.24 & 32.88 \\
			\text{SSL-HSIC \cite{li2021self}} & 91.95 & 89.99 & 67.23 & 57.01 & 92.09 & 88.91 & 51.37 & 36.03 \\
			\text{RELIC \cite{mitrovic2020representation}} & 91.96 & 89.35 & 67.24 & 56.88 & 91.15 & 86.21 & 49.17 & 32.97 \\
			\midrule
			\textbf{ICL-MSR(SimCLR + MSR)} & 92.34 & 89.47 & 67.59 & 57.64 & 92.03 & 86.94 & 50.12 & 32.88 \\
			\textbf{ICL-MSR(BYOL + MSR)} & 92.26 & \textbf {90.12} & 66.97 & \textbf {57.97} & \textbf {93.22} & \textbf {89.36} & 52.54 & \textbf {37.54} \\
			\textbf{ICL-MSR(LMCL + MSR)} & \textbf{92.45} & 89.38 & \textbf {67.99} & 57.71 & 91.56 & 87.73 & \textbf {52.61} & 32.35 \\
			\textbf{ICL-MSR(ReSSL + MSR)} & 91.77 & 89.06 & 65.12 & 55.07 & 89.91 & 88.06 & 47.17 & 33.03\\
			\bottomrule
		\end{tabular}
	\end{center}
	\vspace{-0.1cm}
\end{table*}

In the second stage of each epoch, to learn the parameters of $f_{msr}$, we propose a meta learning-based training mechanism. That is, $f_{msr}$ is updated by encouraging the weight matrix to be chosen such that, if $f$ and $f_{ph}$ are trained using the weight matrix, the performance of the primary contrastive learning would be maximized on this same training data. Specifically, We first update $f$ and ${f _{ph}}$ once with the learning rate $\alpha$ by the follows:
\begin{equation}
    \begin{array}{l}
{f ^1} = f  - \alpha {\nabla _f }{L_{to}},\\
f _{ph}^1 = {f _{ph}} - \alpha {\nabla _{{f _{ph}}}}{L_{to}},
\end{array}
\end{equation}
These two updates can be seen as to learn a good $f$ and a good $f_{ph}$. After this step, ${f ^1}$ and ${f _{ph}^1}$ can be seen as the function of ${f _{msr}}$, because ${\nabla _f }{L_{to}}$ and ${\nabla _{{f _{ph}}}}{L_{to}}$ is related to ${f _{msr}}$. Then, we update ${f _{msr}}$ by minimizing the following: 
\begin{equation}
\label{88}
\mathop {\min }\limits_{{f _{msr}}} {L_{ct}}\left( {{f ^1},f _{ph}^1} \right) + \gamma {L_{uni}}
\end{equation}
where $\gamma$ is a hyper-parameter, ${L_{ct}}( {{f ^1},f _{ph}^1})$ represents that the loss $L_{ct}$ is calculated based on the parameters ${f ^1}$ and ${f _{ph}^1}$, and ${L_{uni}}$ is the uniformity loss that aims to constrain the distribution of elements in $A_s$ to approximate a uniform distribution, so that the resulting visual semantics can be as inconsistent as possible. Based on the Gaussian potential kernel \cite{b1, b2, o2}, the ${L_{uni}}$ can be represented as :
\begin{equation}
{L_{uni}} = \log \sum\limits_{{a_i},{a_j} \in {A_s}} {{G_t}\left( {{a_i},{a_j},t} \right)}
\end{equation}
where ${{G_t}\left( {{a_i},{a_j},t} \right) \buildrel \Delta \over = \exp \left( {2t \cdot {a_i}^{\mathop{\rm T}\nolimits} {a_j} - 2t} \right)}$, and $t$ is a fixed hyper-parameter. Note that the average pairwise Gaussian potential is nicely tied with the uniform distribution, for more details please refer to \cite{o2}. 

A problem is that why minimizing objective (\ref{88}) can make $f_{msr}$ to learn semantic information. Note that only the semantic information shared between the positive pair can prompt $X_i^{3 - j}$ and $Y(X_i^{3 - j})$ to be similar, thus minimizing the contrastive loss. Based on the SCM, we can see that the shared semantic information contains both background-related and foreground-related information. The idea behind objective (\ref{88}) is that minimizing the contrastive loss so that $f_{msr}$ learns the shared semantic information. Also, this step can be seen as promoting the learning of ${L_{ct}}$ once again on the basis of ${L_{ct}}$, which is similar to learning to learn. This is also the reason why we call it a meta semantic regularizer.

\section{Error Bound for Downstream Classification}
The classification task is often used to evaluate the performance of most CL methods. Therefore, we present the generalization error bound (GEB) of the proposed ICL-MSR based on the classification task which trains a softmax classifier by minimizing the traditional cross entropy loss \cite{zhang2018generalized}, e.g., ${L_{SM}}( {f;T} ) = {\inf _W}{L_{CE}}( {W \cdot f;T} )$, where $W$ is the linear classifier, $T$ is the label. For a feature embedding
$f(X)$, the generalization error is defined by $L_{SM}^T( f ) = {\mathbb{E}_X}[ {{L_{SM}}( {f;T} )} ]$. Then we investigate how such a generalization error $L_{SM}^T( f )$ is far from the contrastive learning objective $L_{ct}$.

\begin{theorem}
\label{thm:bigtheorem}
Let ${f^*} \in \arg {\min _f}{L_{cl}} + \lambda {L_{msr}}$. Then with probability at least $1 - \delta$, we have that
\begin{equation} \label{qw}
    \left| {L_{SM}^T\left( {{f^*}} \right) - {L_{cl}}\left( {{f^*}} \right)} \right| \le O\left( {\frac{{{Q_1}{\mathscr{R}_H}\left( \lambda  \right)}}{M} + \sqrt {\frac{{{Q_2}}}{M}} } \right)
\end{equation}
where $M$ is the total number of training samples, $N$ is the size the mini-batch, and ${Q_1} = \sqrt {1 + {1 \mathord{\left/
 {\vphantom {1 N}} \right.
 \kern-\nulldelimiterspace} N}} $, ${Q_2} = \log \left( {{1 \mathord{\left/
 {\vphantom {1 \delta }} \right.
 \kern-\nulldelimiterspace} \delta }} \right) \cdot {\log ^2}\left( M \right)$, ${{\mathscr{R}_H}\left( \lambda  \right)}$ is the rademacher complexity. Also, ${{\mathscr{R}_H}\left( \lambda  \right)}$ is monotonically decreasing w.r.t. $\lambda$. 
\end{theorem}

The detailed proof can be found in Appendix \ref{aaa}. As shown in equation (\ref{qw}), we can obtain that the error bound gradually decreases with the increase of the training sample size $M$. Note that this observation is consistent with the traditional supervised learning method. Also, we can see that the mini-batch size $N$ in the error term $\sqrt {{{{Q_2}} \mathord{\left/
 {\vphantom {{{Q_2}} N}} \right.
 \kern-\nulldelimiterspace} N}} $ is negligible for the large sample size $N$. In this case, the relative large size $N$ will effectively reduce the first error term ${{Q_1}{\mathscr{R}_H}\left( \lambda  \right)}$, and thereby tightening the error bound. Finally, when we enlarge the regularization parameter $\lambda$, the Rademacher complexity $\mathscr{R}_H$ will also be decreased, and thus further reducing the error bound and improving the generalizability of the contrastive learning algorithm.

\section{Experiments}

\subsection{Benchmark Datasets}
The following datasets are utilized to evaluate the performance of the proposed ICL-MSR: \textbf{CIFAR-10} and \textbf{CIFAR-100} \cite{krizhevsky2009learning} are two small-scale datasets, which consist of images of size $32 \times 32$ from 10 classes and 100 classes, respectively. \textbf{STL-10} \cite{coates2011analysis} is derived from ImageNet and consists of more than 100K training samples with $96 \times 96$ resolution. \textbf{Tiny ImageNet} \cite{le2015tiny} can be seen as a simplified version of ImageNet, which contains 100K training samples and 10K testing samples
from 200 classes and an image scale of $64 \times 64$. \textbf{ImageNet-100} \cite{tian2020contrastive} is a randomly sampled subset of ImageNet with a total of 100 classes. \textbf{ImageNet} \cite{ deng2009imagenet} is a well-known large-scale dataset. It consists of about 1.3M training images and 50K test images with over 1000 classes.

\begin{figure}[t]
	\centering
	\includegraphics[width=0.48\textwidth]{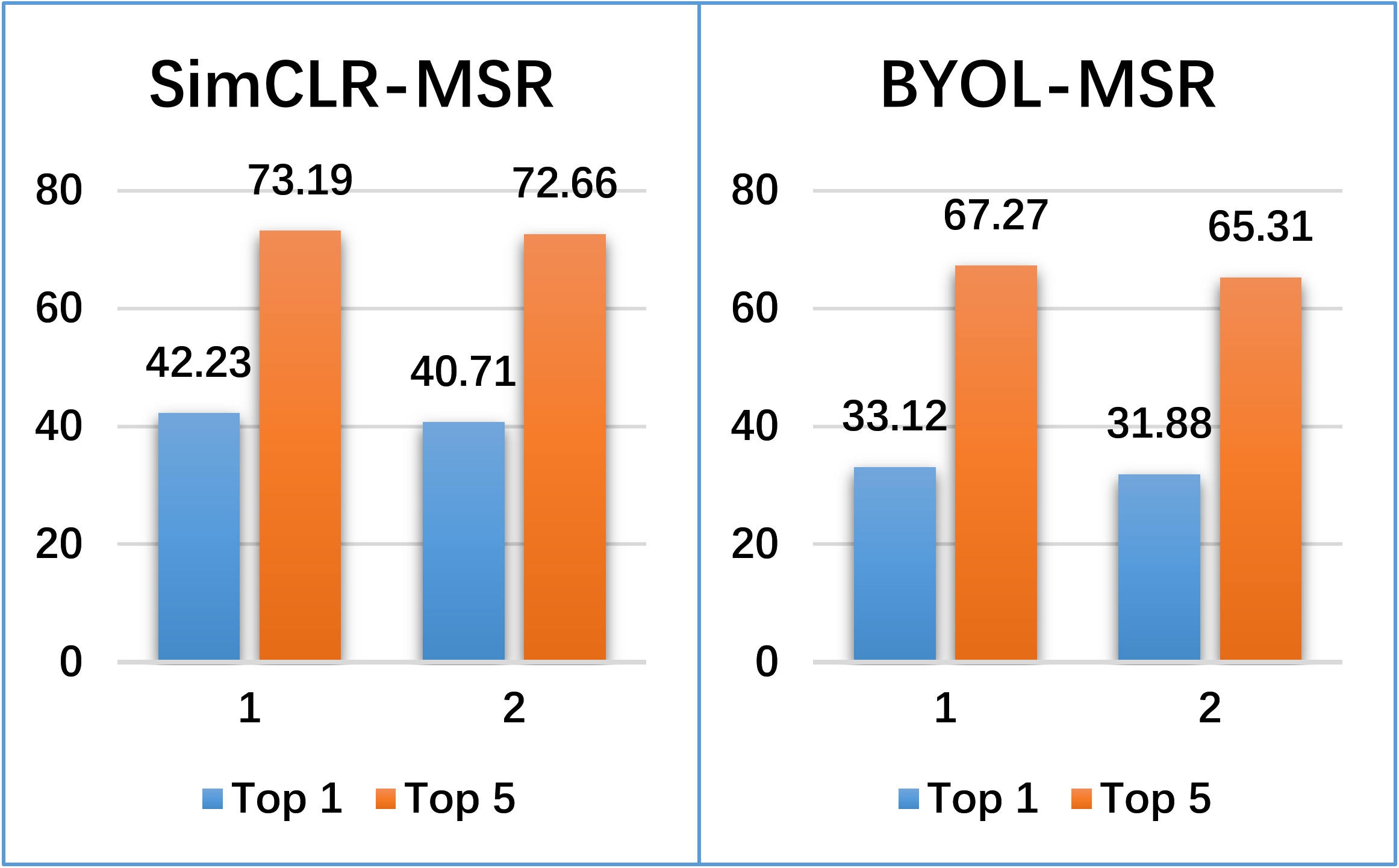}
	\vskip -0.1in
	\caption{The experimental results for two kinds of ICL-MSR models. "1" represents training and testing on full images, "2" represents training on full images and testing on foreground images.}
	\label{555}
	\vskip -0.2in
\end{figure}

\subsection{Implementation Details}
The experiments we carried out are to evaluate the effectiveness of the proposed meta semantic regularizer. So, we ignore the influence of secondary factors, e.g., the neural network architecture. To this end, for CIFAR-10, CIFAR-100, STL-10, and Tiny ImageNet, we use the same feature extractor for all compared methods, e.g., ResNet-18. For ImageNet and ImageNet-100, we use ResNet-50 as the feature extractor. For all datasets, the obtained feature representations are $L_2$ normalized, unless otherwise specified. We set $t = 2$ and $n=6$. For SimCLR, we set $ \tau = 0.5 $. For BYOL, we use the exponential moving average with cosine increasing, starting from 0.99. For all the compared methods, the Adam optimizer \cite{kingma2014adam} is used for the datasets with small and medium sizes. Also, for CIFAR-10 and CIFAR-100, the number of epochs is set to 1,000 and the learning rate is set to $3 \times {10^{ - 3}}$, for Tiny ImageNet, ImageNet-100, the number of epochs is set to 1,000 and the learning rate is set to $2 \times {10^{ - 3}}$, for STL-10, the number of epochs is set to 2,000 and the learning rate is set to $2 \times {10^{ - 3}}$. Also, for all datasets, we use learning rate warm-up for the first 500 iterations of the optimizer, and a 0.2 learning rate drop 50 and 25 epochs before the end. The dimension of output of the projection head $f_ph$ is set to 1024. The weight decay is set to $10^-6$. The output dimension of the $f$ is set to 64 for CIFAR-10 and CIFAR-100, 128 for STL-10 and Tiny ImageNet. Finally, for ImageNet, we set the implementation and hyperparameters to be the same as \cite{ chen2020big, chuang2020debiased}.

As for image transformation details, on CIFAR-10 and CIFAR-100, and Tiny ImageNet, the extracted crops are varied with a random size from 0.2 to 1.0 and a random aspect ratio from 3:4 to 4:3 of the input image. The horizontal mirroring is implemented with a probability of 0.5. The probability of the color jittering with configuration (0.4, 0.4, 0.4, 0.1) is set to 0.8.  The probability of grayscaling is set to 0.1. For ImageNet and ImageNet-100, the crop size is set to be varied from 0.08 to 1.0, we use stronger jittering (0.8; 0.8; 0.8; 0.2) and set the probability of grayscaling to 0.2 and the probability of Gaussian blurring to 0.5. As for evaluation protocol, we first freeze the $f$ after training phase and then train a supervised linear classifier on top of it. The linear classifier is a fully-connected layer followed by softmax. We set the epochs for training the linear classifier to 500 with the Adam optimizer. We also report the accuracy of a k-nearest neighbors classifier ($k$-nn), and set $k=5$.

For toy experiments, we run four kinds of methods on COCO dataset \cite{lin2014microsoft}, including SimCLR, BYOL, SimCLR-MSR, and BYOL-MSR. During training, we eliminated the samples containing multiple targets in the coco dataset to ensure that the samples in the training set are only one target. Meanwhile, we observe that some classes in the coco dataset contain only a few or no samples. Therefore, we only selected 30 of these categories, including: \{'airplane':0,
    'banana':1,
    'bear':2,
    'bed':3,
    'bench':4,
    'bird':5,
    'boat':6,
    'broccoli':7,
    'bus':8,
    'cat':9,
    'clock':10,
    'cow':11,
    'dog':12,
    'elephant':13,
    'fire hydrant':14,
    'giraffe':15,
    'horse':16,
    'motorcycle':17,
    'person':18,
    'pizza':19,
    'scissors':20,
    'sink':21,
    'stop sign':22,
    'teddy bear':23,
    'toilet':24,
    'traffic light':25,
    'train':26,
    'truck':27,
    'vase':28,
    'zebra':29\}. The CoCo dataset provides ground-truth bounding boxes of objects in images, so we take the area inside the bounding box of each image as the foreground, and the area outside the bounding box of each image as the background.

\subsection{Evaluation Results}

\begin{table}[t]
	\small
	\renewcommand\arraystretch{1.22}
	\vskip 0.in
	\caption{Classification accuracy (top 1 and top 5) of a linear classifier for methods with the ResNet-50 feature extractor and negative sample size 4096 on ImageNet-100.}
	\vskip 0.15in
	\label{tab:2}
	\setlength{\tabcolsep}{10pt}
	\begin{center}
		\begin{tabular}{lccc}
			\toprule
			Methods & Top 1 & Top 5 \\ 
			\midrule
			\text{SimCLR \cite{i1}} & 70.15 & 89.75  \\
			\text{MoCo \cite{he2020momentum}} & 72.80 & 91.64 \\ 
			\text{CMC \cite{tian2020contrastive}} & 73.58 & 92.06 \\
			\text{SwAV \cite{caron2020unsupervised}} & 75.78 & 92.86 \\
			\text{DCL \cite{chuang2020debiased}} & 74.60 & 92.08 \\
			\text{LMLC \cite{ermolov2021whitening}} & 75.89 & 92.89 \\
			\midrule
			\textbf{ICL-MSR (SimCLR + MSR)} & 72.08 & 91.81 \\
			\textbf{ICL-MSR (CMC + MSR)} & 74.60 & 92.87  \\
			\textbf{ICL-MSR (CMC + SwAV)} & 75.91 & 92.88  \\
			\textbf{ICL-MSR (DCL + MSR)} & 75.68 & 93.17  \\
			\textbf{ICL-MSR (LMLC + MSR)} & \textbf{76.45} & \textbf{93.88}  \\
			\bottomrule
		\end{tabular}
	\end{center}
	\vspace{-0.2cm}
\end{table}

\begin{table}[t]
	\small
	\renewcommand\arraystretch{1.22}
	\vskip 0.in
	\caption{Classification accuracy (top 1 and top 5) of a linear classifier for methods with the ResNet-50 feature extractor on ImageNet.}
	\vskip 0.15in
	\label{tab:3}
	\setlength{\tabcolsep}{10pt}
	\begin{center}
		\begin{tabular}{lccc}
			\toprule
			Methods & Top 1 & Top 5 \\ 
			\midrule
			\text{SimCLR \cite{i1}} & 69.3 & 89.0  \\
			\text{MoCo \cite{he2020momentum}} & 71.1 & - \\ 
			\text{CMC \cite{tian2020contrastive}} & 66.2 & 87.0 \\
			\text{CPC \cite{henaff2020data}} & 63.8 & 85.3 \\
			\text{InfoMin Aug. \cite{tian2020makes}} & 73.0 & 91.1 \\
			\text{SwAV \cite{caron2020unsupervised}} & 75.3 & - \\
			\text{BYOL \cite{chuang2020debiased}} & 74.3 & 91.6 \\
			\text{RELIC \cite{ermolov2021whitening}} & 74.8 & 92.2 \\
			\text{SSL-HSIC \cite{li2021self}} & 72.2 & 90.7 \\
			\midrule
			\textbf{ICL-MSR (SimCLR + MSR)}  & 70.9 & 90.5 \\
			\textbf{ICL-MSR (SwAV + MSR)}  & \textbf {75.5} & - \\
			\textbf{ICL-MSR (BYOL + MSR)} & 75.4 & \textbf {92.6}  \\
			\bottomrule
		\end{tabular}
	\end{center}
	\vspace{-0.2cm}
\end{table}

\begin{figure*}[t]
	\centering
	\includegraphics[width=1\textwidth]{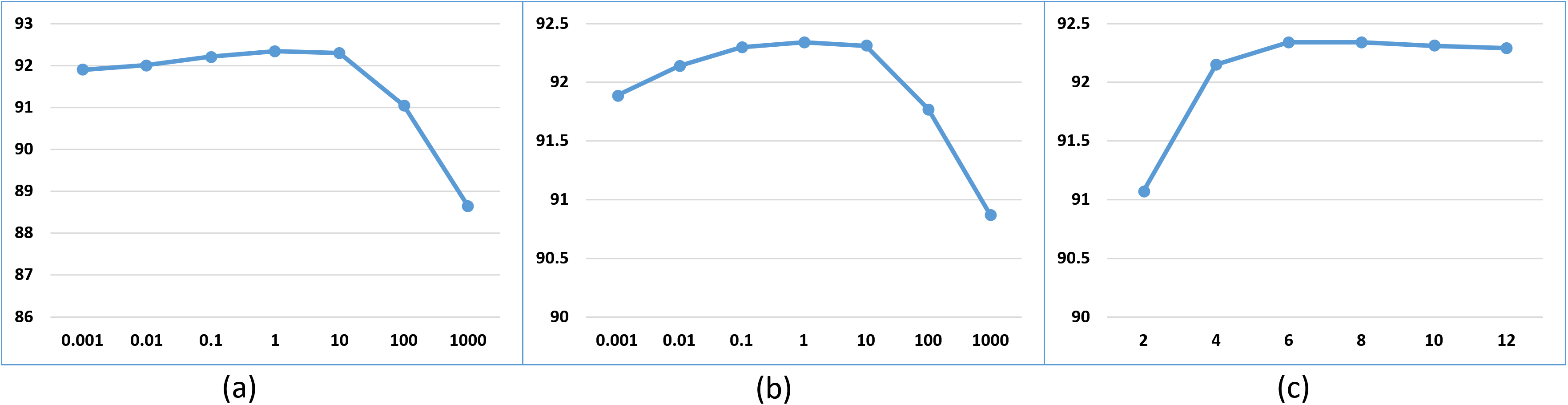}
	\vskip -0.1in
	\caption{Impacts of the hyperparameters (a) $\lambda$, (b) $\gamma$, and (c) the number of semantic weight vectors $n$.}
	\label{pp1}
	\vskip -0.2in
\end{figure*}

Figure \ref{555} shows the experimental results of two kinds of proposed ICL-MSR including SimCLR+MSR and BYOL+MSR. When training ICL-MSR on the full images, we can observe that the testing results on full images are comparable with those on foreground images. Also, the testing results on full images have increased compared with those in Figure \ref{toy}. This indicates that the proposed backdoor adjustment-based regularization method is effective and the background information is indeed a confounding factor that can interference with the learning process of the feature extractor. 

Table \ref{tab:1} shows the experimental results (linear and 5-nn) of the compared methods with a ResNet-18 feature extractor on small and medium size datasets. The compared methods include SimCLR, BYOL, Barlow Twins, W-MSE, ReSSL, LMCL, SSL-HSIC, and RELIC. We incorporate the Meta Semantic Regularizer into four models, resulting in four kinds of ICL-MSR (SimCLR+MSR, BYOL+MSR, LMCL+MSR, ReSSL+MSR). The hyper-parameters $\lambda$ and $\gamma$ are set to $1$ and $1$ for SimCLR+MSR, $0.1$ and $1$ for BYOL+MSR and LMCL+MSR, $0.01$ and $1$ for ReSSL+MSR, respectively. Table \ref{tab:2} shows the experimental results (Top 1 and Top 5) of the compared methods with a ResNet-50 feature extractor on ImageNet-100. The compared methods include SimCLR, MoCo, CMC, SwAV, DCL, and LMLC. We incorporate the Meta Semantic Regularizer into four models, resulting in four kinds of ICL-MSR (SimCLR+MSR, CMC+MSR, DCL+MSR, LMLC+MSR). The hyper-parameters $\lambda$ and $\gamma$ are set to $0.1$ and $1$ for SimCLR+MSR, $0.1$ and $0.1$ for CMC+MSR, and $1$ and $1$ for DCL+MSR and LMLC+MSR, respectively. Table \ref{tab:3} shows the experimental results (Top 1 and Top 5) of the compared methods a ResNet-50 feature extractor on ImageNet. The compared methods include SimCLR, MoCo, CMC, CPC, InfoMin Aug., SwAV, BYOL, SSL-HSIC, and RELIC. We incorporate the Meta Semantic Regularizer into two models, resulting in two kinds of ICL-MSR (SimCLR+MSR, BYOL+MSR). The hyper-parameters $\lambda$ and $\gamma$ are set to $0.1$ and $1$, respectively.

From the three Tables, we can observe that the performances of our proposed ICL-MSR are all better than those of the baseline, For Table \ref{tab:1}, the best results always appear in BYOL+MSR and LMCL+MSR. For Table \ref{tab:2}, the best results are located in LMCL+MSR. For Table \ref{tab:3}, LMCL+MSR outperforms all compared methods. This indicates the effectiveness of the proposed ICL-MSR. Note that RELIC is a causal-related method and focuses on the different augmentations. We can observe that, in Table \ref{tab:1}, three of the four proposed methods outperform RELIC, and in Table \ref{tab:2}, BYOL+MSR outperforms RELIC. This indicates that background information is indeed a confounding factor for learning a good feature extractor. We summarize two possible reasons why the improvement is less than 1\% in most cases. The first is that ICL-MSR is more suitable for image data with larger resolutions. We find that improvement of less than 1\% mostly occurred on datasets with small resolutions, e.g., $32\times32$ for cifar-10 and cifar-100 datsets. An image with a smaller resolution already contains less background information. The second is that ICL-MSR may be sensitive to the hyperparameter $n$. To reduce computational complexity, we set $n=6$ for all datasets.

We evaluate the performance obtained when fine-tuning the representation learned by ICL-MSR on a classification task with a small subset of ImageNet’s training dataset. We follow the semi-supervised protocol of \cite{i1, chuang2020debiased} and use the same fixed splits of respectively 1\% and 10\% of ImageNet labeled training dataset. We report both top-1 and top-5 accuracies on the test dataset in Table \ref{tab:4}. We set $n=14$. We can observe that the proposed ICL-MSR outperforms the compared methods in all cases. This indicates that the proposed ICL-MSR is effective for the fine-turning tasks.

\begin{table}[t]
	%\small
	\renewcommand\arraystretch{1.22}
	\vskip -0.1in
	\caption{Semi-supervised training with a fraction of ImageNet labels (1\% and 10\%). Classification accuracy (top 1 and top 5) of a linear classifier for methods with the ResNet-50 feature extractor.}
	\vskip -0.in
	\label{tab:4}
	\setlength{\tabcolsep}{10pt}
	\begin{center}
	\resizebox{\linewidth}{!}{
		\begin{tabular}{lccccc}
			\toprule
			\multirow{2}{*}{Methods} & \multicolumn{2}{c}{1\%} & \multicolumn{2}{c}{10\%} \\ \cline{2-5}
			& Top 1 & Top 5 & Top 1 & Top 5 \\
			\midrule
			\text{SimCLR \cite{i1}} & 48.3 & 75.5 & 65.6 & 87.8 \\
            \text{BYOL \cite{chuang2020debiased}} & 53.2 & 78.4 & 68.8 & 89.0 \\
            \midrule
            \textbf{ICL-MSR (SimCLR + MSR)} & 50.7 & 77.1 & 66.9 & 89.6 \\
            \textbf{ICL-MSR (BYOL + MSR)} & \textbf {55.5} & \textbf {80.6} & \textbf {70.5} & \textbf {90.7} \\
            \bottomrule
		\end{tabular}}
	\end{center}
	\vspace{-0.2cm}
\end{table}

\subsection{Hyperparametric Analysis}
To understand the impacts of hyper-parameters, we select SimCLR+MSR to conduct comparisons on CIFAR-10 dataset. Specifically, $\lambda$ controls the impact of the MSR, $\gamma$ controls the impact of the uniformity loss, and $n$ is the number of the learned weight vectors. We first set $\gamma = 1, n=6$ and select $\lambda$ from the range of $\{ {{{10}^{ - 3}},{{10}^{ - 2}},...,{{10}^3}} \}$. The results are shown in Figure \ref{pp1} (a). We observe that when $\lambda = 1$, ICL-MSR gets the best accuracy. This illustrates that the proposed MSR is effective. Then, we set $\lambda=1, n=6$ and select $\gamma$ from the range of $\{ {{{10}^{ - 3}},{{10}^{ - 2}},...,{{10}^3}} \}$. From Figure \ref{pp1}(b), we obtain that the best result corresponds to $\gamma = 1$. This indicates that constraining the distribution of the weight matrix to a uniform distribution is effective. Finally, we set $\lambda=\gamma=1$ and select $n$ from the range of $\{ {2,4,...,12} \}$. From Figure \ref{pp1}(c), we can obtain that a proper number of semantic weight vectors can prompt the performance of ICL-MSR. When $\gamma = 10^3$, the accuracy is the lowest. This indicates that releasing uniform constraint will discard the semantic information.

\section{Conclusions}
In this paper, based on the toy experiments on the COCO dataset with four experimental settings, we find two contradictory conclusions. Then, we build a Structural Causal Model (SCM) to give an explanation for this contradiction and propose to regard the background as a confounder. To tackle this problem, we propose a regularization method based on backdoor adjustment. Our method can be easily incorporated into most existing CL methods. We demonstrate the effectiveness of the proposed method both theoretically and empirically.

\section*{Acknowledgements}
The authors would like to thank the anonymous reviewers for their valuable comments. This work is supported in part by National Natural Science Foundation of China No. 61976206 and No. 61832017, Key Special Project for Introduced Talents Team of Southern Marine Science and Engineering Guangdong Laboratory (Guangzhou) No. GML2019ZD0603, National Key Research and Development Program of China No. 2019YFB1405100, Beijing Outstanding Young Scientist Program NO. BJJWZYJH012019100020098, Beijing Academy of Artificial Intelligence (BAAI), the Fundamental Research Funds for the Central Universities, the Research Funds of Renmin University of China 21XNLG05, and Public Computing Cloud, Renmin University of China. This work is also supported in part by Intelligent Social Governance Platform, Major Innovation \& Planning Interdisciplinary Platform for the “Double-First Class” Initiative, Renmin University of China, and Public Policy and Decision-making Research Lab of Renmin University of China.

\section*{Statement}
We find that the positions of $X_s$ and $X^{3-j}_i$ in Figure 2 in the camera-ready version submitted to ICML is wrong. The Figure 2 in the blind-reviewed version is correct. Since the camera-ready deadline has passed, the 2022 Publications Chairs tell us it is impossible to update the camera ready version at this point and there is also no official errata. So, we release the correct version in arXiv. Also, we have carefully checked the whole paper. 

\nocite{langley00}
\bibliography{ICL-MSR}
\bibliographystyle{icml2022}

%%%%%%%%%%%%%%%%%%%%%%%%%%%%%%%%%%%%%%%%%%%%%%%%%%%%%%%%%%%%%%%%%%%%%%%%%%%%%%%
%%%%%%%%%%%%%%%%%%%%%%%%%%%%%%%%%%%%%%%%%%%%%%%%%%%%%%%%%%%%%%%%%%%%%%%%%%%%%%%
% APPENDIX
%%%%%%%%%%%%%%%%%%%%%%%%%%%%%%%%%%%%%%%%%%%%%%%%%%%%%%%%%%%%%%%%%%%%%%%%%%%%%%%
%%%%%%%%%%%%%%%%%%%%%%%%%%%%%%%%%%%%%%%%%%%%%%%%%%%%%%%%%%%%%%%%%%%%%%%%%%%%%%%
\newpage
\appendix
\onecolumn
\icmltitle{Appendix: Interventional Contrastive Learning with Meta Semantic Regularizer}
\section{The Training Process} 

\begin{algorithm}[H]
\renewcommand{\algorithmicrequire}{\textbf{Input:}}
\renewcommand\algorithmicensure {\textbf{Output:} }
\caption{ICL-MSR}
\label{alg:algorithm}

\begin{algorithmic}[1]

\REQUIRE ~~\\ %算法的输入参数：Input
\text{\#}: $N$, ~minibatch size \\
\text{\#}: $f$, ~encoder function \\
\text{\#}: $f_{msr}$, ~meta semantic regularizer \\
\text{\#}: $f_{ph}$, ~projection head \\
\text{\#}: $\lambda$, $\gamma$, $n$, $t$, $\tau $, ~hyperparameters \\
\text{\#}: $\alpha$, $\beta$, ~learning rates \\

%\ENSURE ~~\\ %算法的输出：Output
%\text{\#}: encoder function $f(\cdot)$\\
\REPEAT
\FOR{$t$-th training iteration}
\STATE Iteratively sample minibatch ${X_{tr}} = \left\{ {{X_i}} \right\}_{i = 1}^N$.
\STATE $\# \ regular \ contrastive \ training \ step$
\STATE $f \leftarrow f - \alpha{\nabla _f}{L_{to}}$ \\
\STATE ${f_{ph}} \leftarrow {f_{ph}} - \alpha{\nabla _{{f_{ph}}}}{L_{to}}$ \\
\ENDFOR
\FOR{$t$-th training iteration}
\STATE Iteratively sample minibatch ${X_{tr}} = \left\{ {{X_i}} \right\}_{i = 1}^N$.
\STATE $\# \ compute \ fast \ weights$
\STATE $\# \ retain \ computational \ graph$
\STATE $f^1 \leftarrow f - \alpha{\nabla _f}{L_{to}}$ \\
\STATE ${f_{ph}^1} \leftarrow {f_{ph}} - \alpha{\nabla _{{f_{ph}}}}{L_{to}}$ \\
\STATE $\# \ meta \ training \ step \ using \ second \ derivative$ \\
\STATE ${f_{msr}} \leftarrow {f_{msr}} - \beta {\nabla _{{f_{msr}}}}\left[ {{L_{ct}}\left( {{f^1},f_{ph}^1} \right) + \gamma {L_{uni}}} \right]$ \\
\ENDFOR
\UNTIL $f$, $f_{ph}$, and $f_{msr}$ converge.
\end{algorithmic}
\end{algorithm}

\section{Derivation of Equation \ref{bd}}
\label{bdbd}
We first give definitions to the path, the d-separation, and the backdoor criterion. From \cite{glymour2016causal}, we can obtain that: 
\begin{definition}
\label{def:1}
\textbf{Path}. A path consists of three components including the Chain Structure: $A \to B \to C$ or $A \leftarrow B \leftarrow C$, the Bifurcate Structure: $A \leftarrow B \to C$, and the Collisions Structure: $A \to B \leftarrow C$.
\end{definition}

\begin{definition}
\label{def:1}
\textbf{$d$-separation}. A path $p$ is blocked by a set of nodes $Z$ if and only if:

1. $p$ contains a chain of nodes $A \to B \to C$ or a fork $A \leftarrow B \leftarrow C$ such that the middle node $B$ is in $Z$ (i.e., $B$ is conditioned on), or

2. $p$ contains a collider $A \to B \leftarrow C$ such that the collision node $B$ is not in $Z$, and no descendant of $B$ is in $Z$. If $Z$ blocks every path between two nodes $X$ and $Y$,  then $X$ and $Y$ are $d$-separated, conditional on $Z$, and thus are independent conditional on $Z$.
\end{definition}

\begin{definition}
\label{def:1}
\textbf{The Backdoor Criterion}. Given on ordered pair of variables $\left( {X,Y} \right)$ in a directed acyclic graph $G$, a set of variables $Z$ satisfies the backdoor criterion relative to $\left( {X,Y} \right)$ if no node in $Z$ is a descendant of $X$, and $Z$ blocks every path between $X$ and $Y$ that contains an arrow into $X$. If a set of variables of $Z$ satisfies the backdoor criterion for $X$ and $Y$, then the causal effect of $X$ on $Y$ is given by the formula:
\begin{equation}
    P\left( {Y = y\left| {do\left( {X = x} \right)} \right.} \right) = \sum\limits_z {P\left( {Y = y\left| {X = x,Z = z} \right.} \right)} P\left( {Z = z} \right)
\end{equation}
\end{definition}

\section{Proof for Theorem \ref{thm:bigtheorem}}
\label{aaa}
First, we give a Lemma as follows:

\begin{lemma}
\label{f:1}
\cite{arora2019theoretical} Assume that ${f^*} \in \arg {\min _f}{L_{cl}} + \lambda {L_{msr}}$. Then with probability at least $1 - \delta$ over the training data $X = \left\{ {{X_1},{X_2},...,{X_M}} \right\}$, for any $f \in \mathcal{H}$,
\begin{equation} \label{qw}
L_{SM}^T\left( {{f^*}} \right) \le {L_{cl}}\left( {{f^*}} \right) + O\left( {\frac{{{Q_1}{\mathscr{R}_H}\left( \lambda  \right)}}{M} + \sqrt {\frac{{{Q_2}}}{M}} } \right)
\end{equation}
where $N$ is the size of negative pairs, ${Q_1} = \sqrt {1 + {1 \mathord{\left/
 {\vphantom {1 N}} \right.
 \kern-\nulldelimiterspace} M}} ,{Q_2} = \log \left( {{1 \mathord{\left/
 {\vphantom {1 \delta }} \right.
 \kern-\nulldelimiterspace} \delta }} \right) \cdot {\log ^2}\left( M \right)$, the Rademacher Complexity is defined as ${\mathscr{R}_H}\left( \lambda  \right) = {E_{\sigma  \in {{\left\{ { \pm 1} \right\}}^{3dN}}}}\left[ {{{\sup }_{f \in H\left( \lambda  \right)}}\left\langle {\sigma ,f} \right\rangle } \right]$, $d$ is the dimension of the learned feature representation, and the restricted hypothesis space is defined as $H\left( \lambda  \right) = \left\{ {\varphi \left| {\varphi  \in H,{\rm{ }}and{{\cal R}_1}\left( f \right) \le {4 \mathord{\left/
 {\vphantom {4 \lambda }} \right.
 \kern-\nulldelimiterspace} \lambda }} \right.} \right\}$.
\end{lemma}

\begin{theorem}
\label{thm:1}
Let ${f^*} \in \arg {\min _f}{L_{cl}} + \lambda {L_{msr}}$. Then with probability at least $1 - \delta$, we have that
\begin{equation} \label{qw}
    \left| {L_{SM}^T\left( {{f^*}} \right) - {L_{cl}}\left( {{f^*}} \right)} \right| \le O\left( {\frac{{{Q_1}{\mathscr{R}_H}\left( \lambda  \right)}}{M} + \sqrt {\frac{{{Q_2}}}{M}} } \right)
\end{equation}
where $M$ is the total number of training samples, $N$ is the negative pair size, and ${Q_1} = \sqrt {1 + {1 \mathord{\left/
 {\vphantom {1 N}} \right.
 \kern-\nulldelimiterspace} N}} $, ${{\mathscr{R}_H}\left( \lambda  \right)}$ is the rademacher complexity. Also, ${{\mathscr{R}_H}\left( \lambda  \right)}$ is monotonically decreasing w.r.t. $\lambda$.
\begin{proof}
 For the traditional cross entropy loss $L_{SM}^T\left( f \right)$, we have that:
\begin{equation}
\begin{array}{l}
L_{SM}^T\left( f \right)\\
 = {\mathbb{E}_X}\left[ {\mathop {\inf }\limits_W {L_{CE}}\left( {W \cdot f,T} \right)} \right]\\
 = {\mathbb{E}_X}\left[ { - \log \frac{{{e^{f{{\left( X \right)}^T}{u_{{c^ + }}}}}}}{{{e^{f{{\left( X \right)}^T}{u_{{c^ + }}}}} + \sum {{e^{f{{\left( X \right)}^T}{u_{{c^ - }}}}}} }}} \right]\\
 = {\mathbb{E}_X}\left[ { - \log \frac{{{e^{f{{\left( X \right)}^T}{\mathbb{E}_{{X^ + }}}\left[ {f\left( {{X^ + }} \right)} \right]}}}}{{{e^{f{{\left( X \right)}^T}{\mathbb{E}_{{X^ + }}}\left[ {f\left( {{X^ + }} \right)} \right]}} + M\mathbb{E}\left[ {{e^{f{{\left( X \right)}^T}{E_{{X^ - }}}\left[ {f\left( {{X^ - }} \right)} \right]}}} \right]}}} \right]\\
 \ge {\mathbb{E}_X}\left[ { - \log \frac{{{e^{f{{\left( X \right)}^T}{\mathbb{E}_{{X^ + }}}\left[ {f\left( {{X^ + }} \right)} \right]}}}}{{{e^{f{{\left( X \right)}^T}{\mathbb{E}_{{X^ + }}}\left[ {f\left( {{X^ + }} \right)} \right]}} + M{\mathbb{E}_{{X^ - }}}\left[ {{e^{f{{\left( X \right)}^T}{\mathbb{E}_{{X^ - }}}\left[ {f\left( {{X^ - }} \right)} \right]}}} \right]}}} \right]\\
 = {L_{cl}}\left( f \right)
\end{array}
\end{equation}
Then, we can obtain:
\begin{equation}
    L_{SM}^T\left( {{f^*}} \right) - {L_{cl}}\left( f \right) \le {L_{cl}}\left( {{f^*}} \right) - {L_{cl}}\left( f \right) \le O\left( {\frac{{{Q_1}{\mathscr{R}_H}\left( \lambda  \right)}}{M} + \sqrt {\frac{{{Q_2}}}{M}} } \right)
\end{equation}
Therefore, we have
\begin{equation} \label{qw}
    \left| {L_{SM}^T\left( {{f^*}} \right) - {L_{cl}}\left( {{f^*}} \right)} \right| \le O\left( {\frac{{{Q_1}{\mathscr{R}_H}\left( \lambda  \right)}}{M} + \sqrt {\frac{{{Q_2}}}{M}} } \right)
\end{equation}

\end{proof}

\end{theorem}

\end{document}